\documentclass[runningheads]{llncs}
\usepackage[T1]{fontenc}
\usepackage{graphicx}
\usepackage{booktabs}
\usepackage{comment}
\usepackage{amsmath}
\usepackage{graphicx}
\usepackage{wrapfig}
\usepackage{amssymb}
\usepackage{booktabs}
\usepackage{multirow}
\usepackage{textcomp}

\usepackage[misc]{ifsym}

\usepackage{mwe}

\begin{document}

\title{Weighted Contrastive Learning for Anomaly-Aware Time-Series Forecasting}


\author{Joel Ekstrand, Tor Mattsson, Zahra Taghiyarrenani, Slawomir Nowaczyk,Jens Lundström, Mikael Lindén}





\institute{Center for Applied Intelligent Systems Research, Halmstad University, Halmstad, Sweden\\
 \email{\{Joel.Ekstrand, zahra.taghiyarrenani, slawomir.Nowaczyk, Jens.Lundström\}@hh.se}
\and Mikael Linden Consulting AB, Stockholm, Sweden \email{Mikael@mikaellindenconsulting.com}}

\maketitle              

\begin{abstract}
Deep forecasters achieve strong accuracy on typical time-series data yet routinely fail in production when demand shifts abruptly, a gap that benchmark evaluations rarely surface but that industrial deployments, such as ATM cash logistics, encounter regularly. In cash management, unexpected demand spikes or dropouts caused by crises, outages, or local disruptions directly translate into service failures or capital inefficiency, making forecasting reliability under distribution shift a hard operational requirement. We identify the core deployment challenge as a stability–sensitivity trade-off: a model must remain accurate under routine variation but must also respond correctly when an anomaly genuinely changes the forecast trajectory. Standard fine-tuning on anomaly-augmented data poorly resolves the trade-off; it recovers anomaly-period accuracy at the cost of catastrophic forgetting on normal data. We propose Weighted Contrastive Anomaly-Aware Adaptation (WECA), a training objective that addresses this trade-off directly by scaling contrastive alignment between normal and anomaly-augmented representations according to perturbation severity. Mild variations are strongly aligned for stability; severe anomalies are only partially aligned, preserving the distributional signal needed for accurate forecasting under shifts. Evaluation on a nationwide ATM transaction dataset (1.3k locations, two years) with domain-calibrated anomaly injection, designed to replicate the high-impact events that cause real operational failures, shows that WECA improves SMAPE on anomaly-affected data by 6.1 percentage points over a normally trained baseline while leaving normal-period accuracy virtually unchanged. Neither fine-tuning nor standard contrastive learning achieves this balance. These results demonstrate that the stability–sensitivity trade-off can be resolved at training time, without architectural changes or post-deployment adaptation, yielding a forecaster that remains reliable across the operating conditions a production deployment actually encounters.

\keywords{Time-Series Forecasting \and Distribution Shift \and Anomaly Robustness \and ATM Cash Demand \and Contrastive Learning}
\end{abstract}

\section{Introduction}
\label{sec:intro}

Accurate forecasting of ATM cash withdrawals is critical for cash logistics and service continuity \cite{rafi2020atm,assaf2025structural}, where forecasting errors directly translate into service outages or capital inefficiency; this gap incurs concrete operational costs. 

ATM networks produce multivariate time series with strong, learnable seasonal structure. Modern deep forecasters exploit this structure effectively, achieving low error on stationary test sets \cite{zhao2026survey}. Yet cash demand is occasionally disrupted by sudden, high-magnitude events, local crises, payment outages, and coordinated withdrawal behavior that induce sharp shifts in distribution. These anomalous situations are rare enough to be underrepresented in training data but severe enough to have a significant operational impact when they occur. The result is a class of deployment failures that standard train/test splits are ill-equipped to expose: a model that appears production-ready on held-out normal data but fails precisely when reliable forecasting matters most.

A natural response to a shift in data after deployment is post-hoc adaptation: fine-tuning the deployed model on new data. Although the model will learn new patterns, they may not be permanent; they could be caused by some anomalous situation, and the situation may return to its normal state. In practice, simply finetuning models leads to catastrophic forgetting, meaning forgetting the normal situation. In other words, the model recovers performance under anomalies at the cost of substantially degraded accuracy during normal operation. 

State-of-the-art deep forecasters (e.g., Transformers, frequency-domain models) \cite{eldele2024label,franceschi2019unsupervised,trirat2024universal,yang2022unsupervised,qiu2026dbloss,wang2026time,wu2022timesnet} perform well on normal data but struggle to maintain reliable performance when the data distribution shifts due to rare or anomalous events. Contrastive learning \cite{hu2024comprehensive,park2024selfsupervised} can improve representation generalization by aligning similar inputs, but strict alignment of normal and anomaly-augmented samples may suppress anomaly signals needed for accurate forecasts in anomalous situations. A method is required that balances stability with sensitivity to anomalous changes, meaning it should forecast accurately during normal periods and still perform well when the data suddenly changes due to anomalies.

We propose Weighted-Contrastive Anomaly-Aware Adaptation (\textbf{WECA}), a training objective that applies weighted contrastive learning to softly align normal and augmented sequences, preserving anomaly-relevant distinctions. Experiments on a large-scale ATM dataset (1.3k ATMs over 2 years) show that WECA improves SMAPE on anomaly-affected data while maintaining accuracy during normal periods.

\textbf{Contributions:} (1) Formalize anomaly-aware forecasting as learning representations that are both invariant to benign variations and sensitive to anomalies. (2) Introduce a weighted contrastive loss that controls alignment strength, enabling a tunable invariance–sensitivity trade-off. (3) Demonstrate improved accuracy under distribution shifts on real-world cash-demand forecasting data without degrading normal performance.

\section{Related Work}
\label{sec:related}

Time-series forecasting has evolved from classical models to deep neural architectures capable of capturing nonlinear and long-range dependencies. RNNs improved sequential modeling but suffered from vanishing gradients and slow recurrence \cite{kim2025comprehensive}.
These limitations motivated the adoption of Transformer-based forecasters, which use self-attention to efficiently model long-horizon dependencies \cite{vaswani2023attention,wang2026deep}. 
Informer \cite{zhou2021informer}, and Autoformer \cite{wu2021autoformer} further reduced the cost of attention, making large-scale forecasting feasible.  
More recently, frequency-domain models such as TimesNet \cite{wu2022timesnet} and FEDformer \cite{zhou2022fedformer} leverage Fourier transforms to extract seasonal patterns with high efficiency, while foundation models like MOMENT \cite{goswami2024moment} pre-train general encoders for transfer learning across datasets.  
In parallel, self-supervised approaches such as TS2Vec \cite{yue2022ts2vec}, SoftCLT \cite{lee2024soft}, and Timesurl \cite{liu2024timesurl} learn robust representations.

Despite these architectural advances, forecasting models trained on normal data often struggle with distribution shifts caused by rare or anomalous events. A straightforward approach involves fine-tuning pre-trained forecasters on anomaly-augmented data, but this risks significant degradation in the model's performance during normal data periods \cite{wu2024adaptive,wang2024comprehensive}. Representation learning provides a principled way to learn invariant features that generalize across changing conditions \cite{trirat2024universal}.
Contrastive learning \cite{hu2024comprehensive} learns robust representations by aligning positive pairs in the latent space, making features invariant to distribution shifts.
In forecasting, however, anomalies often change future values, so forcing full alignment between normal and anomaly-augmented samples can remove useful anomaly information.

We propose to control how strongly pairs are aligned.
Mild perturbations are strongly aligned, while severe anomalies are only partially aligned, preserving anomaly-relevant signals.
This leads to representations that remain stable for normal data but still adapt when severe distribution shifts occur.
\section{Method}
\label{sec:method}

Our goal is to train a forecaster that remains accurate under normal operating conditions while also performing well when the data distribution shifts due to anomalous events. 
Instead of adapting the model after anomalies are observed, we design a training objective that encourages the encoder to learn representations that are both consistent under benign variations and sensitive to anomaly-relevant changes from the start.

Formally, let $\mathbf{x} \in \mathbb{R}^{T \times C}$ denote an input window of $T$ timesteps with $C$ variables, and $\mathbf{y} \in \mathbb{R}^{H \times C}$ the $H$-step forecast horizon. 
A forecasting model consists of an encoder $g_\phi: \mathbb{R}^{T \times C} \to \mathbb{R}^{T' \times D}$, which maps the input into $T'$ latent representations of dimension $D$, and a decoder $h_\psi: \mathbb{R}^{T' \times D} \to \mathbb{R}^{H \times C}$ that produces the forecast $\hat{\mathbf{y}} = h_\psi(g_\phi(\mathbf{x}))$. 

\subsection{Proposed Approach}

We propose \textbf{Weighted Contrastive Anomaly-Aware Forecasting (WECA)}, a weighted contrastive objective that aligns representations of normal and anomaly-augmented inputs while preserving anomaly-specific information. 
Unlike standard contrastive learning, which enforces full invariance, WECA controls the strength of alignment to avoid collapsing meaningful deviations.

Let $z_{i,t}$ and $\tilde{z}_{i,t}$ denote the encoder representations of the original input $\mathbf{x}_i$ and its anomaly-augmented counterpart $\tilde{\mathbf{x}}_i$ at timestep $t$. 
A similarity weight $w^{(i,t)} \in [0,1]$ scales each pair’s contribution: for benign perturbations $w^{(i,t)} \approx 1$ enforces strong similarity, while for severe anomalies $w^{(i,t)}$ is reduced, preserving anomaly-relevant differences.

We implement this alignment using an InfoNCE-style objective:
\begin{align}
A_{i,t} &= \exp\!\big(z_{i,t} \cdot \tilde{z}_{i,t}\big), \\
\mathcal{N}_{i,t} &= 
\sum_{j=1}^{B} \big( \exp\!\big(z_{i,t} \cdot \tilde{z}_{j,t}\big) 
+ \mathbb{1}_{[i \neq j]}\exp\!\big(z_{i,t} \cdot z_{j,t}\big) \big),
\end{align}
where $A_{i,t}$ is the similarity of the positive pair and $\mathcal{N}_{i,t}$ aggregates similarities with all negatives in the batch. 
The WECA loss is:
\begin{equation}
\ell_{\mathrm{WECA}}^{(i,t)} = 
-\, w^{(i,t)} \log \frac{A_{i,t}}{\mathcal{N}_{i,t}}.
\label{eq:WECA}
\end{equation}

\textbf{Forecasting Loss and Joint Objective.}
Predictive accuracy is ensured by minimizing the mean absolute error (MAE):
\[
\mathcal{L}_{\mathrm{forecast}} =
\frac{1}{H}\sum_{t=1}^{H}\lVert y_t - \hat{y}_t \rVert_1,
\]
where $y_t$ and $\hat{y}_t$ are the ground-truth and predicted values. 
The overall training objective combines forecasting and representation learning:
\begin{equation}
\mathcal{L} = 
\frac{1}{B}\sum_{i=1}^{B} \mathcal{L}_{\mathrm{forecast}}^{(i)}
+ \lambda \frac{1}{B T'}
\sum_{i=1}^{B}\sum_{t=1}^{T'}
\ell_{\mathrm{WECA}}^{(i,t)},
\label{eq:joint}
\end{equation}
where $\lambda$ controls the relative strength of the contrastive term.

\textbf{Connection to Instance-Wise Contrastive Learning.}
When $w^{(i,t)}=1$ for all pairs, WECA reduces to instance-level contrastive learning as used in TS2Vec\cite{yue2022ts2vec}. 
This strict alignment may suppress anomaly-relevant signals; WECA avoids this by weighting alignment, enabling the model to remain sensitive to distribution shifts while preserving performance on normal data.

\section{Experiments}
\label{sec:experiments}
We evaluate on a nationwide ATM transaction dataset covering $\sim$1.3k locations over two years, with daily withdrawal amounts as the target variable. We adopt a rolling-origin protocol: $70\%$ training, $10\%$ validation, $20\%$ test. 

\begin{figure}
    \centering
    \includegraphics[width=0.75\linewidth]{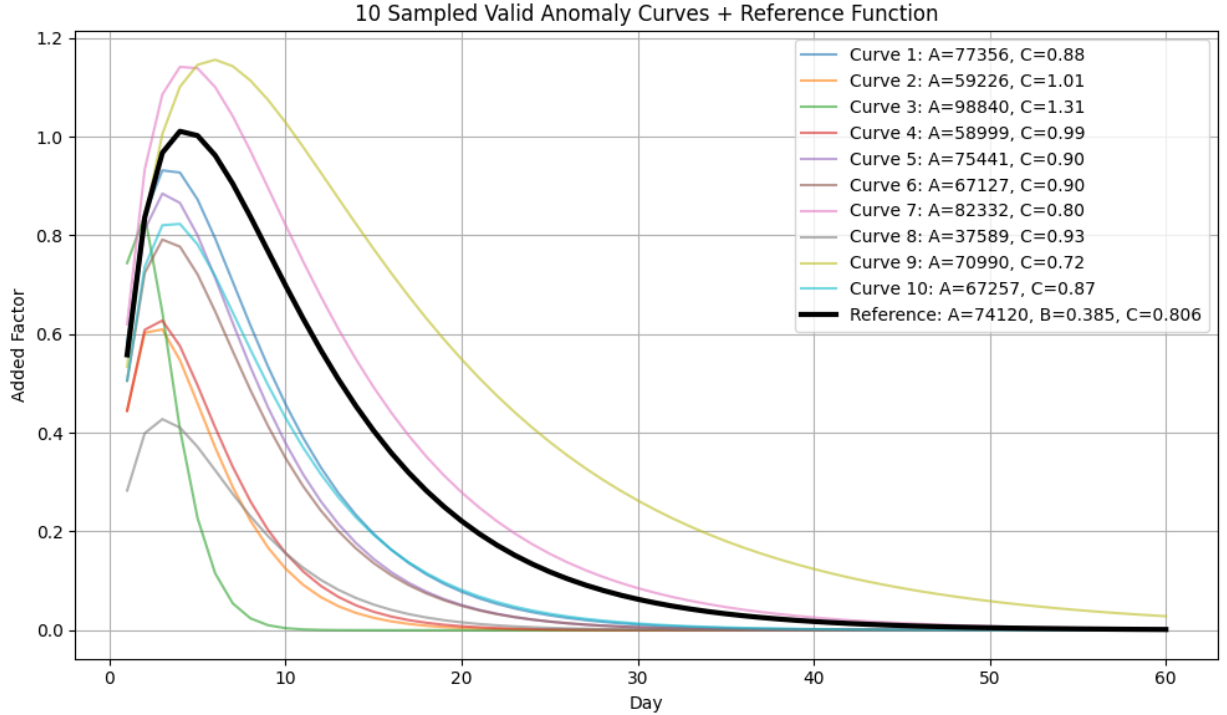}
    \caption{Example of anomaly curves for augmentation. B=0.39 while A and C are sampled from normal distributions with means 74120 and 0.806, and standard deviations 20000 and 0.3.}
    \label{fig:anomaly_examples}
    \includegraphics[width=0.75\linewidth]{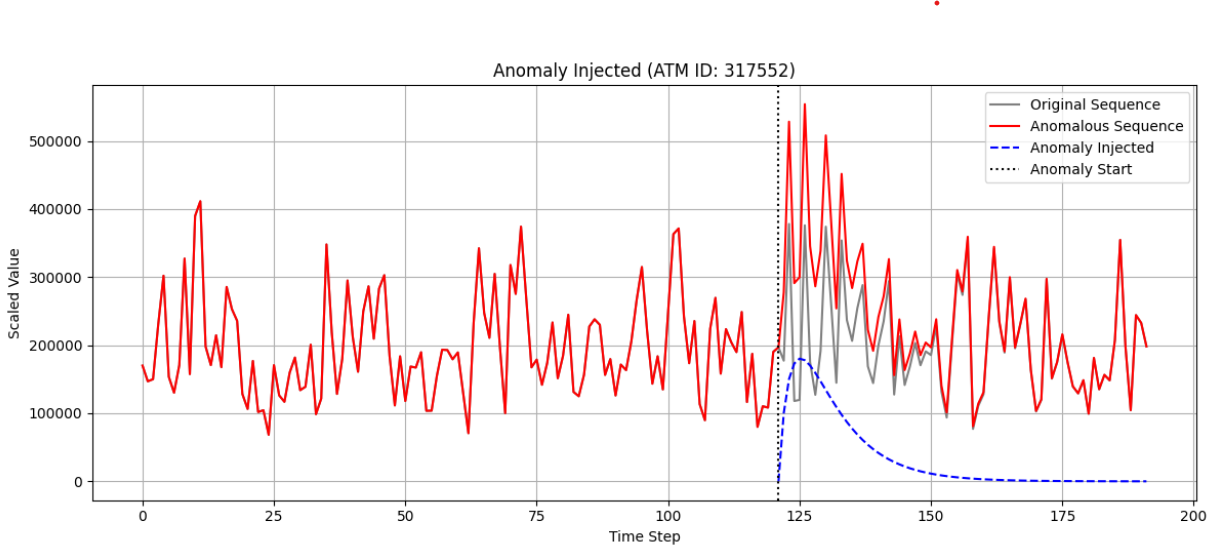}
    \caption{The red graph is the sequence with the injected anomaly (blue dashed graph), added to the grey graph (the original sequence).}
    \label{fig:example_innjected_anomaly}
\end{figure}

\textbf{Synthetic Anomaly Augmentation.}
We do augmentation by creating anomalous samples based on the knowledge of domain experts. 
Each anomaly is parameterized by amplitude, duration, and decay, sampled from distributions fitted to a historical high-impact event. 
The anomaly function is:
$a(n) = \frac{A \cdot n \cdot e^{-B n^{C}}}{90409},$ where $n$ is the day index relative to anomaly onset, and $(A,B,C)$ control severity and shape. 
The anomaly is added to the tail of the input window and propagated into the forecast horizon, so the model observes the onset and must adjust its trajectory. 
This setup produces domain-plausible demand spikes and dropouts and allows reproducible stress-testing of models.
Figures \ref{fig:anomaly_examples} and \ref{fig:example_innjected_anomaly} show one example of augmenting data to present an anomalous situation.

\textbf{Training Details.}  
All models are trained using the Adam optimizer with a learning rate of $10^{-3}$, batch size $B=128$, and early stopping based on validation loss.  
For WECA, we set the weighting coefficient $\lambda=1$, though it can be tuned for different tasks.  
The similarity weights $w^{(i,t)}$ are computed from the Euclidean distance between the original and augmented samples in input space, but the formulation is general and can incorporate other similarity metrics.  
Performance is evaluated using Symmetric Mean Absolute Percentage Error (SMAPE), averaged across all series.

\subsection{Results}
\begin{figure}
    \centering
    \includegraphics[width=0.9\linewidth]{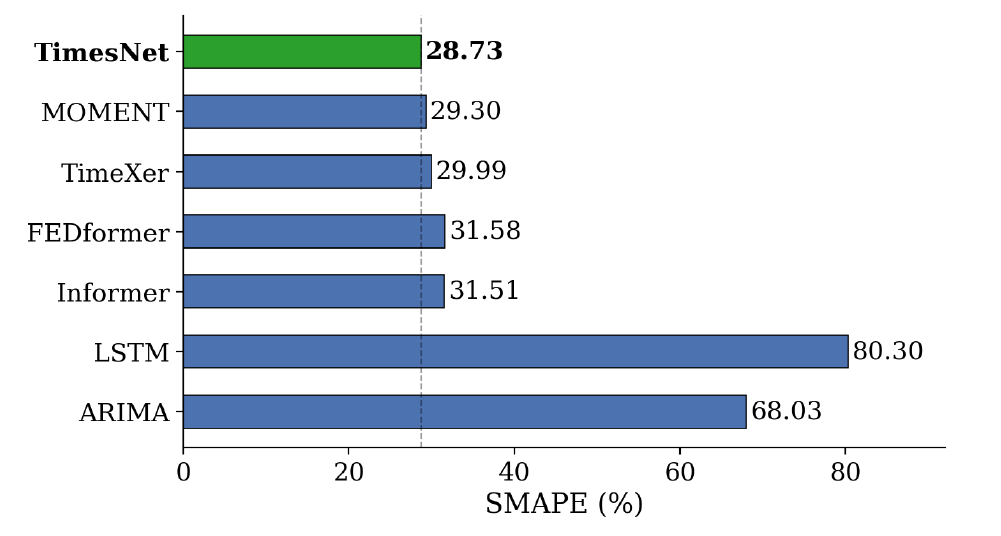}
    \caption{Benchmark SMAPE (\%) on daily transaction data (14-day horizon). TimesNet achieves the lowest error and is used as the backbone.}
    \label{tab:first_results}
\end{figure}

\begin{figure}[h]
    \centering
    \includegraphics[width=0.8\linewidth]{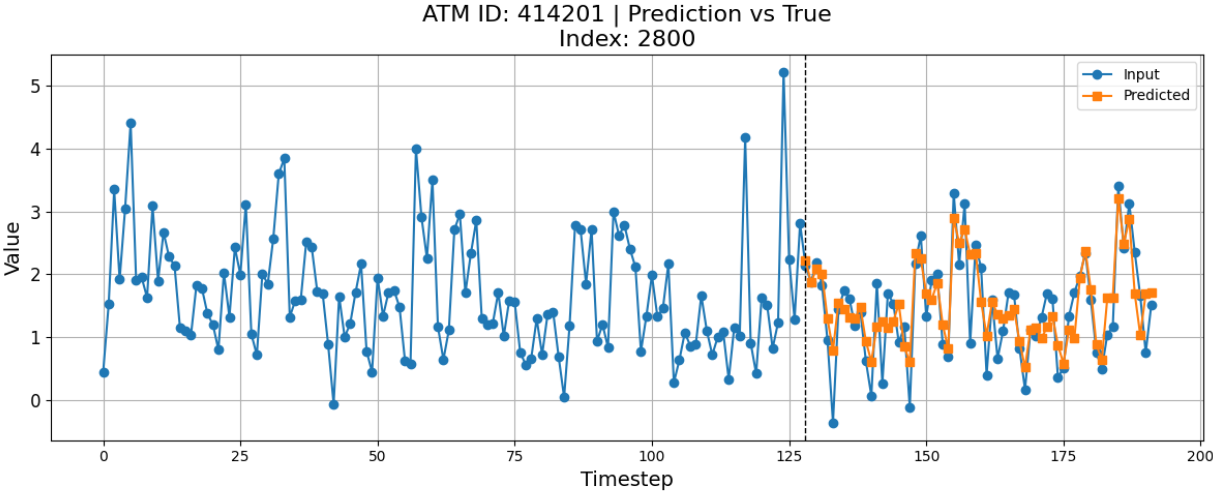}
    \caption{Representative ATM withdrawal time series. 
The blue line shows the historical input sequence; the orange line shows the TimesNet forecast over the prediction horizon. The series exhibits the irregular, high-variance demand patterns typical of the nationwide dataset, underscoring the difficulty of reliable forecasting under real-world operating conditions.}
    \label{fig:timesnet}
\end{figure}

\textbf{Backbone Comparison.}
We benchmark several forecasters on normal data to identify a strong 
backbone for subsequent experiments. As shown in Figure~\ref{tab:first_results}, TimesNet 
achieves the lowest SMAPE of 28.73\% on the 14-day forecasting horizon 
and is used as the backbone for all WECA experiments. Classical models 
(ARIMA, LSTM) lag substantially behind deep forecasters, while 
Transformer-based models (Informer, FEDformer) and recent architectures 
(TimeXer, MOMENT) cluster in the 29--32\% range. Figure~\ref{fig:timesnet} illustrates 
a representative ATM withdrawal series from our dataset, highlighting the irregular, high-variance demand patterns that make this a challenging forecasting task in practice. This figure also shows the results of the TimesNet.

\textbf{Instance vs.\ Temporal Contrastive Loss.}
WECA builds on instance-level contrastive learning, but contrastive objectives for time series can operate along two axes: across instances in the batch (instance loss, IL) or across timesteps within a sequence (temporal loss, TL). Before introducing the WECA weighting, we isolate which contrastive axis provides the stronger inductive bias for anomaly-aware forecasting.

Instance-level contrastive learning (IL) treats each sequence as a unit, pulling together representations of the same sequence under 
different augmentations while pushing apart representations of different sequences. This encourages the encoder to capture global, sequence-level structure, the kind of signal most relevant when an anomaly reshapes the overall demand trajectory. Temporal contrastive learning (TL), by contrast, aligns representations at the level of individual timesteps, encouraging local temporal consistency. While useful for tasks requiring fine-grained temporal detail, this per-timestep alignment risks over-smoothing the encoder representations precisely at the anomaly onset, where abrupt changes are most informative.

\begin{figure}
    \centering
    \includegraphics[width=0.75\linewidth]{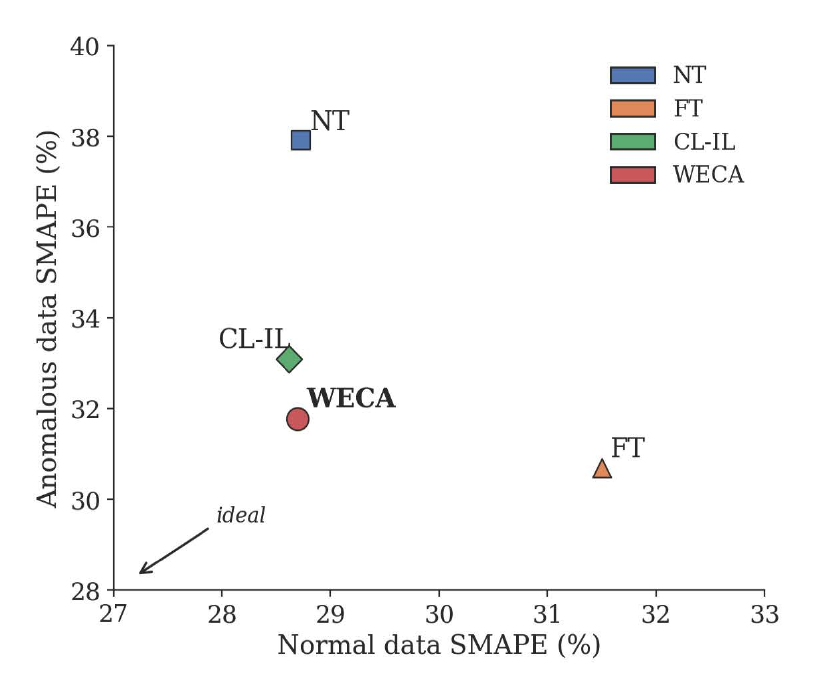}
    \caption{Comparison of contrastive objectives with TimesNet. SMAPE (\%) on normal (ND) and anomalous (AD) data.}
    \label{tab:contrastive_ablation}
\end{figure}

Table~\ref{tab:contrastive_ablation} confirms this intuition. IL achieves the lowest SMAPE on both normal data (28.37\%) and 
anomaly-affected data (31.82\%), outperforming TL and their combination (IL+TL) on both regimes. Combining the two losses (IL+TL) does not 
improve over IL alone; the temporal objective appears to dilute the sequence-level alignment signal rather than complement it. These results establish IL as the stronger basis for contrastive pretraining in this setting and motivate its use as the foundation for the WECA weighting 
scheme.

\begin{table}[h]
\centering
\caption{
Forecasting performance on normal data (ND) and anomaly-affected data (AD), 
reported as SMAPE (\%, mean $\pm$ standard deviation over 5 runs). 
$\Delta$ denotes the change relative to the Normally Trained (NT) baseline 
(lower is better).
}
\label{tab:main_results}

\begin{tabular}{lcccc}
\toprule
\multirow{2}{*}{\textbf{Method}} 
& \multicolumn{2}{c}{\textbf{ND}} 
& \multicolumn{2}{c}{\textbf{AD}} \\
\cmidrule(lr){2-3}
\cmidrule(lr){4-5}
& SMAPE$\downarrow$ & $\Delta$ 
& SMAPE$\downarrow$ & $\Delta$ \\
\midrule
NT 
& 28.73 
& -- 
& 37.91 
& -- \\

FT 
& $31.50 \pm 0.87$ 
& +2.77 
& $30.69 \pm 2.03$ 
& -7.22 \\

CL-IL 
& $28.62 \pm 0.97$ 
& -0.11 
& $33.09 \pm 0.96$ 
& -4.82 \\

\textbf{WECA} 
& $28.70 \pm 1.00$ 
& \textbf{-0.03} 
& $31.78 \pm 1.93$ 
& \textbf{-6.13} \\

\bottomrule
\end{tabular}
\end{table}

\textbf{WECA Results.} We compare WECA against three baselines:  
(i) \textit{No Adaptation (NT)}: trained on normal data only,  
(ii) \textit{Fine-Tuning (FT)}: adapted on anomalouse samples, and  
(iii) \textit{Instance Contrastive Learning (CL-IL)} contrastive pretraining with $w^{(i,t)}=1$.  
We show the results on normal data (ND) and anomalous data (AD).
All models use the same backbone and training setup. 
Table~\ref{tab:main_results} shows that FT achieves the largest AD improvement but severely degrades ND performance, indicating catastrophic forgetting. 
CL-IL improves AD performance moderately without harming ND. 
\textit{WECA achieves the best trade-off by significantly improving performance under anomalies while keeping normal data accuracy nearly unchanged.}
Specifically, WECA reduces SMAPE on AD by 6.13 points compared to the NT, meaning it adapts well to distribution shifts.
At the same time, its SMAPE on ND changes by only 0.03 points, which is practically negligible, so it does not sacrifice accuracy during regular operation.

\section{Conclusion}
\label{sec:conclusion}
We proposed \textbf{Weighted Contrastive Anomaly-Aware Adaptation (WECA)}, a training objective that resolves the stability--sensitivity trade-off inherent in deploying forecasters under the distribution shift. By weighting 
contrastive alignment between normal and anomaly-augmented sequences. According to perturbation severity, WECA learns representations that remain stable under routine variation while preserving the distributional signal needed to forecast accurately when anomalies occur.

Evaluated on a nationwide ATM cash demand dataset with domain-calibrated anomaly injection, WECA improved SMAPE on anomaly-affected data by 6.13 percentage points over a normally trained baseline,  while normal-period accuracy changed by only 0.03 percentage points, a trade-off that neither fine-tuning nor standard contrastive learning achieves. Fine-tuning recovers anomaly-period performance at the cost of catastrophic forgetting; standard contrastive learning moderately improves robustness but incompletely. WECA resolves both failure modes in a single jointly optimized objective, without architectural changes or post-deployment adaptation. These results have direct implications for industrial deployment: anomaly robustness need not be traded off against normal-regime reliability. A forecaster trained with WECA can be deployed with confidence that it will perform well both during regular operations and when demand shifts abruptly, the condition that matters most when forecasting errors translate into 
service outages or capital inefficiency.

\paragraph{Limitations and future work.} 
Anomaly injection in this work is synthetic, calibrated to historical high-impact events but not drawn from observed deployment failures. Validating WECA against a broader set of real anomaly types, including gradual drift, sensor failures, and multivariate co-shifts, remains an open direction. The weighting scheme currently uses Euclidean distance in input space; learned or task-adaptive weights may further improve the invariance--sensitivity balance. Finally, extending WECA to online settings, where anomaly statistics are unknown at training time, is a natural next step 
toward fully autonomous deployment.

\begin{credits}
\subsubsection{\ackname} This work is partially supported by Vinnova, Sweden.
\end{credits}

\bibliographystyle{splncs04}
\bibliography{refs}

\end{document}